%% file: 00-main.tex
\pdfoutput=1

\documentclass[11pt]{article}

\usepackage[]{acl}

\usepackage{times}
\usepackage{latexsym}

\usepackage[T1]{fontenc}

\usepackage[utf8]{inputenc}

\usepackage{microtype}
\usepackage{booktabs}
\usepackage{multirow}
\usepackage[compact]{titlesec}
\usepackage{caption}
\usepackage{subcaption}
\usepackage{graphicx}

\title{\emph{SocioProbe} \\ What, When, and Where  Language Models Learn about Sociodemographics}
\title{\emph{SocioProbe}: What, When, and Where Language Models Learn \\about Sociodemographics}

  \author{Anne Lauscher\textsuperscript{1}, Federico Bianchi\textsuperscript{2}, Samuel Bowman\textsuperscript{3}, and Dirk Hovy\textsuperscript{4} \\
  \textsuperscript{1}Data Science Group, University of Hamburg, Germany \\
  \textsuperscript{2}StanfordNLP, Stanford University, CA, USA \\
  \textsuperscript{3}New York University, NY, USA \\
  \textsuperscript{4}MilaNLP, Bocconi University, Milan, Italy \\
  \texttt{anne.lauscher@uni-hamburg.de}, \texttt{fede@stanford.edu},  \\ \texttt{bowman@nyu.edu}, \texttt{dirk.hovy@unibocconi.it} \\}

\newcommand{\suite}{{\textsc{SocioProbe}}}
\newcommand{\al}[1]{\textcolor{black}{#1}}

\begin{document}
\maketitle
\begin{abstract}
Pre-trained language models (PLMs) have outperformed other NLP models on a wide range of tasks. %
\al{Opting for a more thorough understanding of their capabilities and inner workings}, researchers have established the extend to which they capture lower-level knowledge like grammaticality, and mid-level semantic knowledge like factual understanding. However, there is still little understanding of their knowledge of  higher-level aspects of language. In particular, despite the importance of sociodemographic aspects in shaping \al{our} language, the questions of whether, where, and how PLMs encode these aspects, e.g., gender or age, is still unexplored. 
We address this research gap by probing the sociodemographic knowledge of \al{different} \al{single-GPU PLMs} on multiple \al{English} data sets via traditional classifier probing and information-theoretic minimum description length probing. Our results show that PLMs do encode these sociodemographics, \al{and that this knowledge is sometimes spread across the layers of some of the tested PLMs}. We further conduct a multilingual analysis and investigate the effect of supplementary training to \al{further} explore to what extent, where, and with what amount of pre-training data \al{the knowledge is encoded}. 
Our overall results indicate that sociodemographic knowledge is still a major challenge for NLP. PLMs require large amounts of pre-training data to acquire the knowledge and models that excel in general language understanding do not seem to own more knowledge about \al{these} aspects.
\end{abstract}

\section{Introduction}
\input{01-intro}

\section{Related Work}
\input{02-rw}

\section{SocioProbe}
\input{03-probing-suite}

\section{Experiments}
\input{04-experiments}

\section{Conclusion}
\input{06-conclusion}

\section*{Acknowledgements}
The work of Anne Lauscher is funded under the Excellence Strategy of the German Federal Government and the Länder. This work is in part funded by the  European Research Council (ERC) under the European Union’s Horizon 2020 research and innovation program (grant agreement No. 949944, INTEGRATOR). At the time of writing, AL, FB, and DH were members of the Data and Marketing Insights unit of the Bocconi Institute for Data Science and Analysis

\section*{Limitations}
\input{07-limitations}

\bibliography{custom}
\bibliographystyle{acl_natbib}

\appendix
\clearpage

\input{xx-appendix}

\end{document}

%% file: 01-intro.tex
When talking to somebody, we consciously choose how to represent ourselves, and we have a mental model of who our conversational partner is. At the same time, our language is littered with subconscious clues about our sociodemographic background that we cannot control (e.g., our age, education, regional origin, social class, etc). People use this information as an integral part of language, to better reach their audience, and to understand what they are saying~\citep[e.g.,][]{trudgill}. In other words, we use sociodemographic knowledge to decide what to say (are we talking to a child or an adult, do I want to sound smart or relatable?)
But do pre-trained language models (PLMs) have knowledge about sociodemographics?

Over the last years, PLMs like BERT~\citep{devlin-etal-2019-bert} and RoBERTa~\citep{liu2019roberta} have achieved superior performance on a wide range of downstream tasks~\citep[e.g.,][\emph{inter alia}]{wang-etal-2018-glue,NEURIPS2019_4496bf24}. Accordingly, they have become the \emph{de facto} standard for most NLP tasks.
Consequently, many researchers have tried to shed light on PLMs' inner workings~\citep[cf. \emph{``Bertology''};][]{tenney-etal-2019-bert,rogers-etal-2020-primer}. They have systematically probed the models' capabilities to unveil which language aspects their internal representations capture. In particular, researchers have probed lower-level structural knowledge~\citep[e.g.,][\emph{inter alia}]{hewitt-manning-2019-structural,sorodoc-etal-2020-probing,chi-etal-2020-finding,pimentel-etal-2020-information}, as well as mid-level knowledge, e.g., lexico-semantic knowledge~\citep[e.g.,][]{vulic-etal-2020-probing,beloucif-biemann-2021-probing-pre}, and PLMs' factual understanding~\citep[e.g.,][]{petroni2019language,zhong-etal-2021-factual}. 
While these aspects are relatively well explored, we still know little about higher-level knowledge of PLMs: only a few works have attempted to quantify common sense knowledge in the models~\citep[]{petroni2019language,lin-etal-2020-birds}. Probing of other higher-level aspects still remains underexplored -- hindering targeted progress in advancing human-like natural language understanding.

As recently pointed out by \citet{hovy-yang-2021-importance}, sociodemographic aspects play a central role in language. However, they remain underexplored in NLP, despite promising initial findings \cite[e.g.,][]{volkova-etal-2013-exploring,hovy-2015-demographic,lynn-etal-2017-human}.
Importantly, we are not aware of \textit{any} research assessing sociodemographic knowledge in PLMs. This lack is extremely surprising given the availability of resources, and the importance of these factors in truly understanding language.
\paragraph{Contributions.} 
Acknowledging the importance of sociodemographic factors in language, we address a research gap by proposing \suite{}, a novel perspective of probing PLMs for sociodemographic aspects. We demonstrate our approach along two dimensions, (binary) \emph{gender} and \emph{age}, using two established data sets, \al{and with different widely-used easily-downloadable PLMs that can be run on a single GPU}. To ensure validity of our findings, we combine  ``traditional'' classifier probing~\citep[][]{petroni2019language} and information-theoretic minimum distance length (MDL) probing~\citep{voita-titov-2020-information}. Our experiments allow us to answer a series of research questions. We find that PLMs \textit{do} represent sociodemographic knowledge, but that it is acquired in the later stages. This knowledge is also decoupled from overall performance: some models that excel in general language understanding do still not have more knowledge about sociodemographics encoded. We hope that this work inspires more research on the social aspects of NLP.  \al{Our research code is publicly available at \url{https://github.com/MilaNLProc/socio-probe}}.

\section{Research Questions}
We pose five research questions (\textbf{RQs}):

\vspace{0.3em}
\noindent\textbf{RQ1: \emph{To what extend do current PLMs encode sociodemographic knowledge?}} 
Do these models ``know'' about the existence and impact of sociodemographic aspects like age or gender on downstream tasks, as repeatedly shown \citep[e.g.,][]{volkova-etal-2013-exploring,hovy-2015-demographic,benton-etal-2017-multitask}? 
We probe different versions of the RoBERTa~\citep{liu2019roberta} and DeBERTa~\citep{he2021deberta,he2021debertav3} model families. Our findings reveal the varying extent to which sociodemographic knowledge is encoded in different textual domains. Surprisingly, the superior performance of the DeBERTa model on general NLU tasks is not reflected in the encoding of sociodemographic knowledge.

\vspace{0.3em}
\noindent\textbf{RQ2: \emph{How much pre-training data is needed to acquire sociodemographic knowledge?}}
Are sociodemographic aspects present in any data sample, or are they only learned with sufficient amounts of data?
Inspired by \citet{zhang-etal-2021-need}, we use a suite of MiniBERTas~\citep{warstadt-etal-2020-learning} and RoBERTa \emph{base} trained on different amounts of data (1M to 30B). Our results show that sociodemographic knowledge is learned much more slowly than syntactic knowledge and the gains do not seem to flatten with more training data. This indicates that large data portions are needed to acquire sociodemographic knowledge.

\vspace{0.3em}
\noindent\textbf{RQ3: \emph{Where is sociodemographic knowledge located in the models?}}
Sociodemographic aspects influence a wide range of NLP tasks, both at a grammatical level (e.g., part-of-speech tagging  \citet{garimella-etal-2019-womens}) and at a pragmatic level (e.g., machine translation \citet{hovy-etal-2020-sound,saunders-byrne-2020-reducing}). But where do these factors reside themselves in the model?
By probing different layers of the PLM with \suite{}, we find that sociodemographic knowledge is located in the higher layers of most PLMs. This finding is in-line with the intuition that higher-level semantic knowledge is encoded in higher layers of the models~\citep[e.g.,][]{tenney-etal-2019-bert}. However, on some data sets, some of the models show an opposite trend and the differences across the layers seem much less pronounced than for a lower-level control task, in which we predict linguistic acceptability.

\vspace{0.3em}
\noindent\textbf{RQ4: \emph{Does the localization of sociodemographic knowledge in multilingual models differ?}}
Different languages provide different linguistic ways of expressing sociodemographic (and other) aspects: some lexically, some syntactically \cite{johannsen-etal-2015-cross}. Do PLMs that have been exposed to multiple languages store sociodemographic knowledge differently than monolingual models?
We probe multilingual models and demonstrate that the results are in-line with the findings from RQ3. Thus, the localization of the sociodemographic knowledge in the multilingual versions does not seem to differ from their monolingual counterparts.

\vspace{0.3em}
\noindent\textbf{RQ5: \emph{What is the effect of different supplementary training tasks on the knowledge encoded in the PLMs' features?}}
\citet{phang2018sentence} demonstrated that through supplementary training on intermediate-labeled tasks (STILTs), the performance for downstream tasks can be improved. We hypothesize that such sequential knowledge transfer can activate sociodemographics in PLMs, as these aspects can act as useful signals, e.g., for sentiment analysis~\citep{hovy-2015-demographic}. However, our experiments show that 
specifically the sociodemographic knowledge in the last layers of the models is overwritten through our STILTs procedures.

\vspace{0.3em}
Overall, the encoding of sociodemographic knowledge is still a major challenge for NLP: \textbf{models that excel in NLU do not have more knowledge about sociodemographics, learning curves do not flatten with more pretraining data, the knowledge is much less located than for other tasks, and learning from other tasks is difficult}. %

%% file: 02-rw.tex
\paragraph{Probing PLMs.} The success of large PLMs has led to researchers developing a range of methods~\citep[e.g.,][]{hewitt-liang-2019-designing,torroba-hennigen-etal-2020-intrinsic} and data sets~\citep[e.g.,][]{warstadt-etal-2020-learning,hartmann-etal-2021-multilingual} for obtaining a better understanding of PLMs. In turn, those approaches also challenge these paradigms~\citep[e.g.,][]{pimentel-etal-2020-information,ravichander-etal-2021-probing}. The most straightforward probing approach relies on training classifiers~\citep[e.g.,][]{petroni2019language} to probe models' knowledge. In contrast, other probing mechanisms are substractive \cite[]{cao-etal-2021-low}, intrinsic~\citep{torroba-hennigen-etal-2020-intrinsic}, or rely on control tasks~\citep{hewitt-liang-2019-designing}. A popular family is 
information theoretic probing~\citep[e.g.,][]{pimentel-etal-2020-information, pimentel-cotterell-2021-bayesian}, like minimum description length (MDL) probing~\citep{voita-titov-2020-information}. We use MDL complementarily to traditional probing to further substantiate our claims. 
Most authors have focused on probing English language models~\citep[e.g.,][\emph{inter alia}]{conneau-etal-2018-cram,liu-etal-2021-probing-across,wu-xiong-2020-probing, koto-etal-2021-discourse}, but some have moved into the multilingual space~\citep[e.g.,][]{ravishankar-etal-2019-probing, kurfali-ostling-2021-probing, shapiro-etal-2021-multilabel-approach}, or probed  multimodal models~\citep[e.g.,][]{prasad-jyothi-2020-accents,hendricks-nematzadeh-2021-probing}.

Researchers have used probing to understand whether PLMs encode knowledge about several aspects of language, and to which extent: researchers have probed PLMs for syntactic knowledge~\citep[e.g.,][]{hewitt-manning-2019-structural, sorodoc-etal-2020-probing}, lexical semantics~\citep{vulic-etal-2020-probing, beloucif-biemann-2021-probing-pre}, factual knowledge~\citep{heinzerling-inui-2021-language, petroni2019language, zhong-etal-2021-factual}, and common sense aspects~\citep{lin-etal-2020-birds} or domain-specific knowledge~\citep{jin-etal-2019-probing,pandit-hou-2021-probing,wu-xiong-2020-probing}. Despite this plethora of works, the sociodemographic knowledge remains underexplored.

\paragraph{NLP and Sociodemographic Aspects.} 
Our language use varies depending on the characteristics of the sender and receiver(s), e.g., their age and gender~\cite{eckert2013language,hovy-yang-2021-importance}. 
Accordingly, researchers in NLP have explored these variations~\citep{rosenthal-mckeown-2011-age, blodgett-etal-2016-demographic} and showed that sociodemographic factors influence model performance~\citep[e.g.,][]{volkova-etal-2013-exploring,hovy-2015-demographic}.
Since then, many researchers have argued that such factors should be taken into account for human-centered NLP~\citep{flek-2020-returning}, and showed gains from sociodemographic adaptation~\citep[e.g.,][]{lynn-etal-2017-human, yang-eisenstein-2017-overcoming,li-etal-2018-towards}. 

Other researchers have exploited the tie between language and demographics to profile authors from their texts~\citep{burger-etal-2011-discriminating, nguyen-etal-2014-gender, ljubesic-etal-2017-language, martinc-pollak-2018-reusable}. In this work, we do not develop methods to predict demographic aspects, but use this task as a proxy to how well sociodemographic knowledge is encoded in our models. 
Another line of research has worked on detecting and removing unfair stereotypical bias towards demographic groups from PLMs~\citep{blodgett2020, shah-etal-2020-predictive}, e.g., gender bias~\citep{may-etal-2019-measuring,lauscher-glavas-2019-consistently,webster2020measuring,lauscher2021sustainable}.  
Most recently and closest to our work, \citet{zhang-etal-2021-sociolectal}\footnote{Note that their interest is in linguistically determined language varieties of social groups, i.e., sociolects, whereas we focus on the interplay between individual \textit{demographic} aspects that go across language varieties: we can express gender independent of whether we speak in dialect or standard language.} investigate the sociodemographic bias of PLMs. They compare the PLMs cloze predictions with answers given by crowd workers belonging to different sociodemographic groups. However, they do not provide further insights of the nature of this knowledge nor how when and where it is encoded. 
Our work unequivocally establishes that PLMs contain sociodemographic knowledge, and shows how it is likely acquired, and where it resides.

%% file: 03-probing-suite.tex
\setlength{\tabcolsep}{9pt}
\begin{table*}[t!]
    \centering
    \small{
    \begin{tabular}{llllrc}
    \toprule
        \textbf{Dataset Name} & \textbf{Textual Domain} & \textbf{Dimension} & \textbf{Label} & \textbf{\# Instances}& \textbf{\% Instances} \\
        \midrule
        \multirow{4}{*}{\textbf{Trustpilot}}
         & \multirow{4}{*}{Product Reviews} 
         & \multirow{2}{*}{Gender} & \emph{Man} & \multirow{2}{*}{5349} & 49.97 \\
         &  &  & \emph{Woman} & & 50.03\\ 
         \cmidrule(lr){3-6}
         & & \multirow{2}{*}{Age} & \emph{Young} & \multirow{2}{*}{5269} & 52.19 \\
         & & & \emph{Old} & & 47.80 \\
         \cmidrule(lr){1-6}
         \multirow{8}{*}{\textbf{RTGender}}  & \multirow{2}{*}{Facebook Posts (Congress Members)} & \multirow{2}{*}{Gender} & \emph{Man} & \multirow{2}{*}{510135}& 75.16 \\
         & & & \emph{Woman} & & 24.84\\
         \cmidrule(lr){2-6}
          & \multirow{2}{*}{Facebook Posts (Public Figures)} & \multirow{2}{*}{Gender} & \emph{Man} & \multirow{2}{*}{133,017}&  33.38\\
         & & & \emph{Woman} & & 66.62\\
         \cmidrule(lr){2-6}
        & \multirow{2}{*}{Fitocracy Posts} & \multirow{2}{*}{Gender} & \emph{Man} & \multirow{2}{*}{318,535}&  54.54\\
         & & & \emph{Woman} & & 45.46\\
         \cmidrule(lr){2-6}
        & \multirow{2}{*}{Reddit Posts} & \multirow{2}{*}{Gender} & \emph{Man} & \multirow{2}{*}{1,453,512}&  79.02\\
         & & & \emph{Woman} & & 20,98\\
         \bottomrule
    \end{tabular}}
    \caption{Datasets with dimensions, number of instances (\# Instances), and label distributions (\% Instances).}
    \label{tab:data}
\end{table*}
We describe \suite{}, \al{which we employ to explore the sociodemographic knowledge PLMs contain.} \al{Guided by the availability of data sets, we focus on the dimensions of \emph{gender} and \emph{age}. Note, however, that our overall methodology can be easily extended to other sociodemographic aspects.} 

\subsection{Data}
\label{sec:data}

\begin{figure}[t]
     \centering
     \begin{subfigure}[b]{0.49\textwidth}
         \centering
         \includegraphics[width=\textwidth]{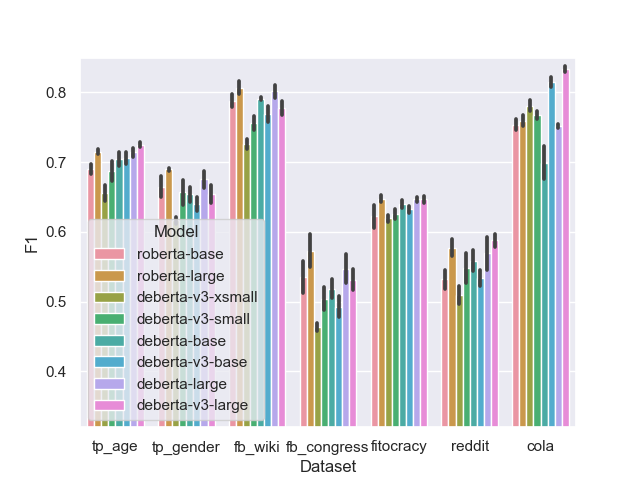}
         \caption{Classic probing}
         \label{fig:all_classic}
     \end{subfigure}
     \begin{subfigure}[b]{0.49\textwidth}
         \centering
         \includegraphics[width=\textwidth]{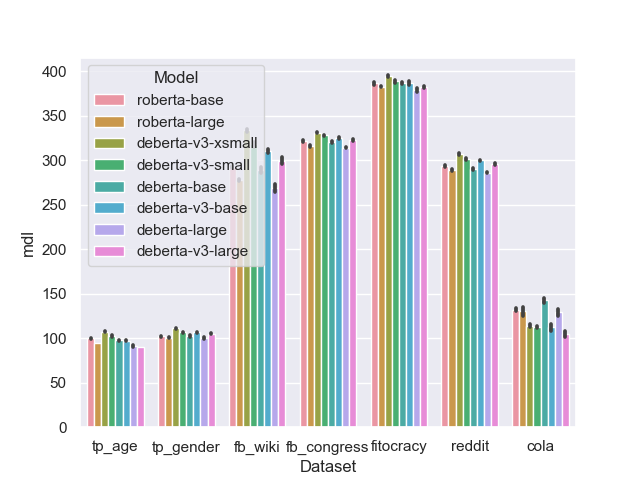}
         \caption{MDL probing}
         \label{fig:all_online}
     \end{subfigure}
    \caption{Results for RQ1. We compare different RoBERTa and DeBERTa models (RoBERTa \emph{base}, RoBERTa \emph{large}, DeBERTa \emph{base}, DeBERTa \emph{large}, DeBERTa v3 \emph{xsmall}, \emph{small}, \emph{base}, and \emph{large}) for (a) classic and (b) MDL probing. We report average and standard deviation of the F1-scores for 5 runs across 7 tasks.}
    \label{fig:all}
\end{figure}
We probe sociodemographic aspects on two data sets. They vary in terms of text length and domain. %

\paragraph{Trustpilot~\citep{hovy-2015-demographic}.} Trustpilot\footnote{\url{https://www.trustpilot.com}} is an international user review platform. The data consists of the review texts (including the rating, which we do not use in this work), as well as the self-identified gender and age of the author. Following the original paper, we do not consider users from 35 to 45 to reduce possible errors due to noisy boundaries. We use the split introduced by~\newcite{hovy-etal-2020-sound}, and focus on the English portion of the data set. For age, we use \textit{Young} for users under the age of 35, and \textit{Old} for people above the age of 45.

\paragraph{RTGender~\citep{voigt-etal-2018-rtgender}.} 
We use all texts of the data set from three different social media platforms: Reddit,\footnote{\url{https://www.reddit.com}} Facebook (posts from politicians and public figures),\footnote{\url{https://www.facebook.com}} and Fitocracy.\footnote{\url{https://www.fitocracy.com}} Our true label corresponds to the gender of the author. In total, the data set consists of 2,415,199 instances. For our experiments, we subsample 20,000 samples for each domain to start from equally-sized portions.

\subsection{Probing Methodology}
We combine two probing methodologies: traditional classifier probing and MDL probing.

\paragraph{Traditional Classifier Probing.} The traditional approach to PLM probing is to place a simple classifier -- the probe -- on top of the frozen features~\citep[e.g.,][\emph{inter alia}]{ettinger-etal-2016-probing, adi2016fine}. In our case, following \citet{tenney-etal-2019-bert} and \citet{zhang-etal-2021-need}, we use a simple two-layer feed-forward network (with rectified linear unit as the activation function) with a softmax output layer. We feed it the average hidden representations of the PLM's Transformer. We take care to only average over the representations of the text and ignore special tokens. We report the F1 measure.

\paragraph{Minimum Description Length Probing.} Traditional classifier probing has been criticized for its reliance on the complexity of the probe~\citep{hewitt-liang-2019-designing, voita-titov-2020-information}. To ensure validity of our results, we thus combine classifier probing with an information theoretic approach. Concretely, we use MDL~\cite{voita-titov-2020-information}. The intuition behind MDL is that the more information is encoded, the less data is needed to describe the labels given the representations. As in the implementation of the \emph{online code estimation setting}, we partition the data into 11 non-overlapping portions representing 0\%, 0.1\%, 0.2\%, 0.4\%, 0.8\%, 1.6\%, 3.2\%, 6.25\%,
12.5\%, 25\%, 50\%,  and 100\% of the full data sets with $t$ numbers of examples each: $\{(x_j , y_j )\}_{j=t_{i-1}+1}^{t_i}$ for $1 \le i \le 11$. Next, we train a classifier on each portion $i$ and compute the Loss $\mathcal{L}$ on the next portion $i+1$. The codelength corresponds to the sum of the resulting 10 losses plus the codelength of the first data portion:
\vspace{-0.3em}

\small{
\begin{equation}
    \textnormal{MDL} = t_1 \log_2 2 - \sum_{i=1}^{10} \mathcal{L}_{i+1}\,,
\end{equation}}
\normalsize

\noindent with $t_1$ as the number of training examples in the first portion of the data set. A lower MDL value indicates more expressive features.

%% file: 04-experiments.tex
We describe our experiments. %

\subsection{General Experimental Setup}
All our experiments follow roughly the same experimental setup: For the Trustpilot data, we use the standard splits provided in~\newcite{hovy-2015-demographic}. On all other data sets, described in Section~\ref{sec:data} we apply a standard split, with 80\% of the data for training, 10\% for validation, and 10\% for testing the models. We train all our models in batches of 32 with a learning rate of 1e-3 using the Adam optimizer~\citep{AdamW} (using default parameters from pytorch). We apply early stopping based on the validation set loss with a patience of 5 epochs. If the loss does not improve for an epoch, we reduce the learning rate by 50 \%. We conduct all experiments 5 times with different random initializations of the probes and report the mean and the standard deviation of the performance scores. For all models, we use versions available on Huggingface and we provide links to all models and code bases used in the Supplementary Materials.

\subsection{RQ1: To what extend do PLMs encode sociodemographic knowledge?}
As initial base experiment, we want to establish how well sociodemographic knowledge can be predicted from the features of different PLMs.

\paragraph{Approach.} We test the features extracted from  RoBERTa~\citep{liu2019roberta} in \emph{base} and \emph{large} configuration in comparison to DeBERTa~\citep{he2021deberta} in \emph{xsmall}, \emph{small}, \emph{base}, and \emph{large} configuration. For DeBERTa, different versions are available in the Huggingface repository. We use the original model as well as the v3 version~\citep{he2021debertav3} of \emph{base} and \emph{large}. The v3 employs ELECTRA-style pre-training with gradient disentangled embedding sharing~\citep{clark2020electra} leading to improvements across all GLUE tasks.\footnote{\url{https://github.com/microsoft/DeBERTa\#fine-tuning-on-nlu-tasks}}

\paragraph{Results.} Figure~\ref{fig:all} shows the results. Generally, the trends in the different models are consistent across the two different probing approaches \al{(classic probing and MDL probing)}. Therefore, we conclude the validity of our approach. \emph{The difficulty of the different data sets varies}: the ``easiest'' task is our control task CoLA, in which we probe lower-level syntactic knowledge. The next-easiest task is to predict the gender in Facebook posts of public figures (\emph{fb\_wiki}, e.g., 80.68 \% average F1 score for RoBERTa \emph{large}). In contrast,  predicting the gender of Facebook posts of congress members is relatively difficult for the models (\emph{fb\_congress}, e.g., 57.32 \% average F1 score). This is in line with the findings of \citet{voigt-etal-2018-rtgender}: depending on the domain of text, the sociodemographic aspects of authors are reflected to varying degrees (here: less so in more formal settings). 
Interestingly, we can not confirm the overall superiority of the DeBERTa models. While the DeBERTa v3 \al{\emph{base} and \emph{large}} models outperform RoBERTa on CoLA by a large margin \citep[6.93 percentage points difference between RoBERTa large and DeBERTa large, as per][]{he2021debertav3}, \emph{RoBERTa large seems to encode sociodemographic knowledge similarly well as DeBERTa large}, or to an even larger extent. The same observation holds when comparing DeBERTa versions. This finding warrants further investigation into how different training regimes affect the encoding of higher-level knowledge.

\subsection{RQ2: How much pre-training data is needed to acquire sociodemographic knowledge?}
We test models trained on varying amounts of data.

\paragraph{Approach.} We use the suite of MiniBERTas~\cite{warstadt-etal-2020-learning}, $12$ RoBERTa-like models, which have been trained on 1M, 10M, 100M, and 1B words, respectively.\footnote{Publicly available at \url{https://huggingface.co/nyu-mll/roberta-med-small-1M-1}} 
The data was randomly sampled from a corpus similar to the original BERT~\citep{devlin-etal-2019-bert}. Pretraining data consisted of the English Wikipedia and Smashwords, which is similar to the BookCorpus~\citep{zhu2015aligning}. The size of the model trained on the smallest portion (1M) is \emph{medium small} (6 layers,	8 attention heads, hidden size of 512). The other models were trained with the \emph{base} configuration (12 layers, 12 attention heads, hidden size of 768). For each size, 3 checkpoints are available (the ones which yielded lowest validation perplexity), trained with different hyperparameters. In comparison, we probe the original RoBERTa in \emph{base} configuration~\citep{liu2019roberta}, trained on approximately 30B words. %

\begin{figure}[t]
     \centering
    \begin{subfigure}[b]{0.49\textwidth}
         \centering
         \includegraphics[width=\linewidth]{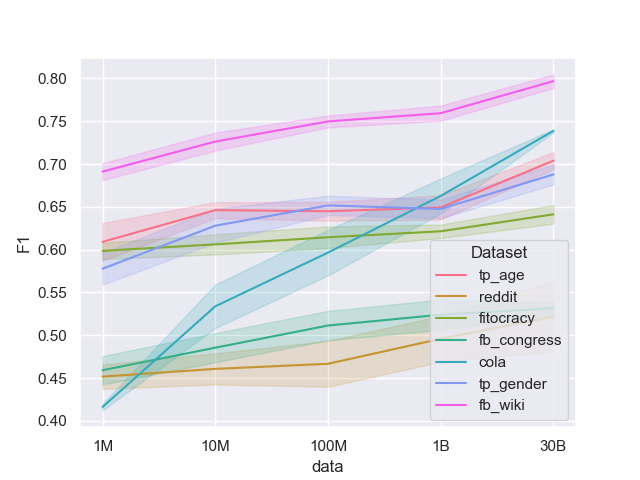}
         \caption{Classic probing}
         \label{fig:miniberta_classic}
     \end{subfigure}
     \hfill
     \begin{subfigure}[b]{0.49\textwidth}
         \centering
         \includegraphics[width=\textwidth]{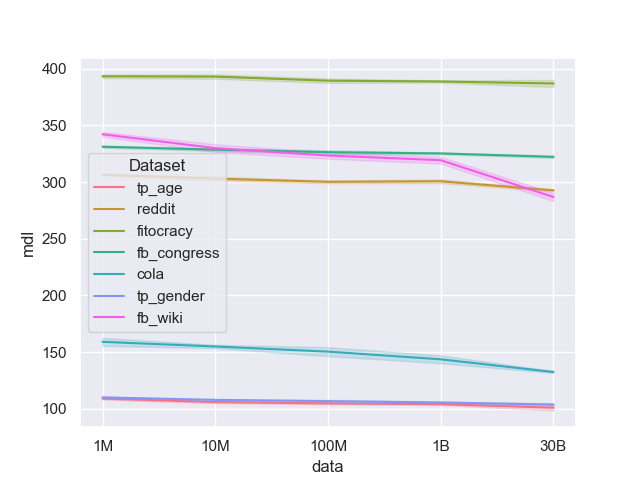}
         \caption{MDL probing}
         \label{fig:miniberta_online}
     \end{subfigure}
    \caption{\al{Classic and MDL probing results for RoBERTa models trained on varying amounts (1M--30B words) of pre-training data (RQ2).}}
\end{figure}
\setlength{\tabcolsep}{11pt}
\begin{table}[t]
    \centering
    \small{
    \begin{tabular}{r r r r r}
    \toprule
         \textbf{\# Tokens} &  \textbf{Costs (\$)} & \textbf{CO$_2$ (lbs)} & \textbf{$\mu_\textnormal{Gain}$} \\
    \midrule
         1M & 50 & 5.825 & -- \\
         10M & 500 & 58.250 & +2.61\\
         100M & 5,075 & 582.500 & +1.98\\
         1B & 20,320 & 2,330.000 & +0.30\\
         30B & 609,600 & 69,900.000 & +8.56\\
     \bottomrule
    \end{tabular}}
    \caption{Results of our cost-benefit analysis. We show financial costs (Costs (\$)) and CO$_2$ emissions (CO$_2$ (lbs)), gain is average F1-measure increase over the ext smaller model across all data sets and models ($\mu_\textnormal{Gain}$).}
    \label{tab:costs}
\end{table}
\begin{figure*}[th!]
     \centering
     \begin{subfigure}[b]{0.248\textwidth}
         \centering
         \includegraphics[width=\textwidth, trim=2em 0.3em 2em 0.3em, clip]{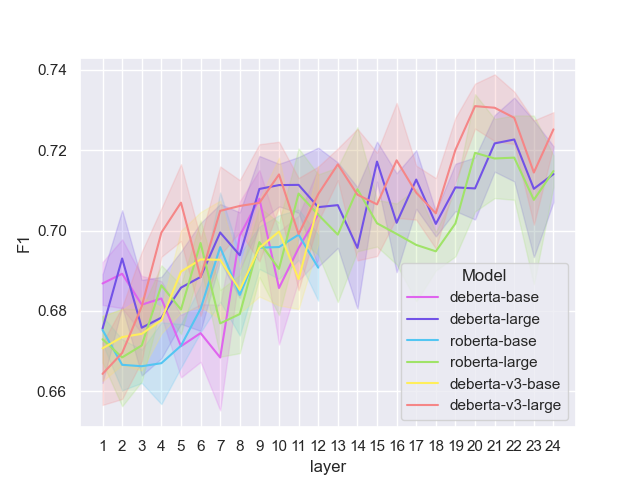}
         \caption{Trustpilot (Age)}
         \label{fig:layers_tp_age}
     \end{subfigure}
     \begin{subfigure}[b]{0.248\textwidth}
         \centering
         \includegraphics[width=\textwidth, trim=2em 0.3em 2em 0.3em, clip]{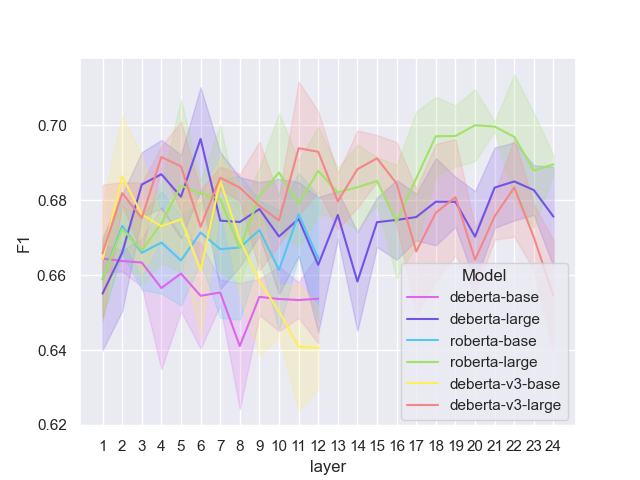}
         \caption{Trustpilot (Gender)}
         \label{fig:layers_tp_gender}
     \end{subfigure}
          \begin{subfigure}[b]{0.248\textwidth}
         \centering
         \includegraphics[width=\textwidth, trim=2em 0.3em 2em 0.3em, clip]{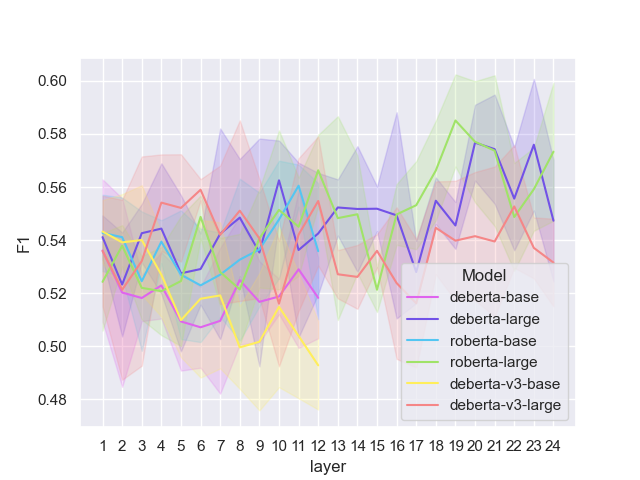}
         \caption{Facebook Congress}
         \label{fig:layers_congress}
     \end{subfigure}
          \begin{subfigure}[b]{0.245\textwidth}
         \centering
         \includegraphics[width=\textwidth, trim=2em 0.3em 2em 0.3em, clip]{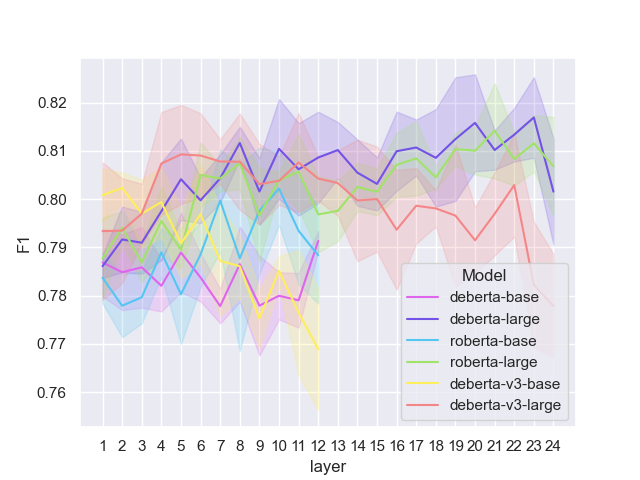}
         \caption{Facebook Wiki}
         \label{fig:layers_wiki}
     \end{subfigure}
          \begin{subfigure}[b]{0.245\textwidth}
         \centering
         \includegraphics[width=\textwidth, trim=2em 0.3em 2em 0.3em, clip]{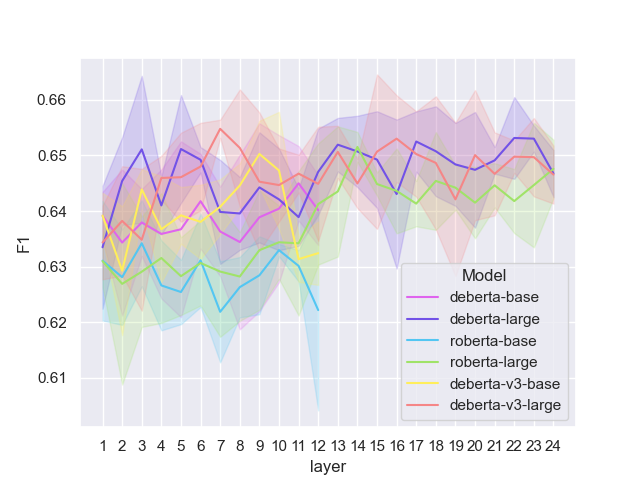}
         \caption{Fitocracy}
         \label{fig:layers_fitocracy}
     \end{subfigure}
          \begin{subfigure}[b]{0.245\textwidth}
         \centering
         \includegraphics[width=\textwidth, trim=2em 0.3em 2em 0.3em, clip]{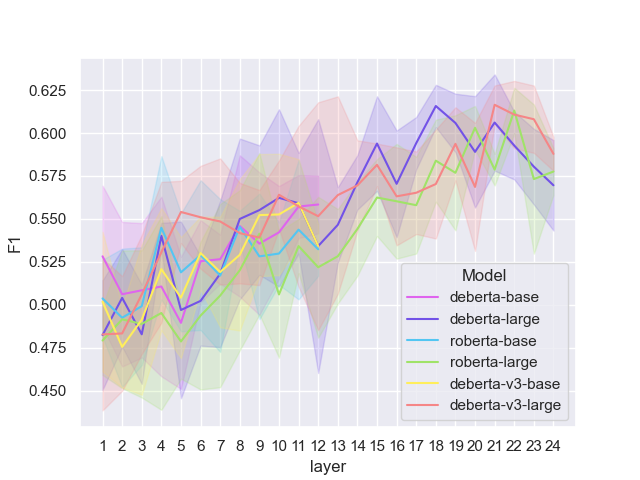}
         \caption{Reddit}
         \label{fig:layers_reddit}
         
     \end{subfigure}
          \begin{subfigure}[b]{0.245\textwidth}
         \centering
         \includegraphics[width=\textwidth, trim=2em 0.3em 2em 0.3em, clip]{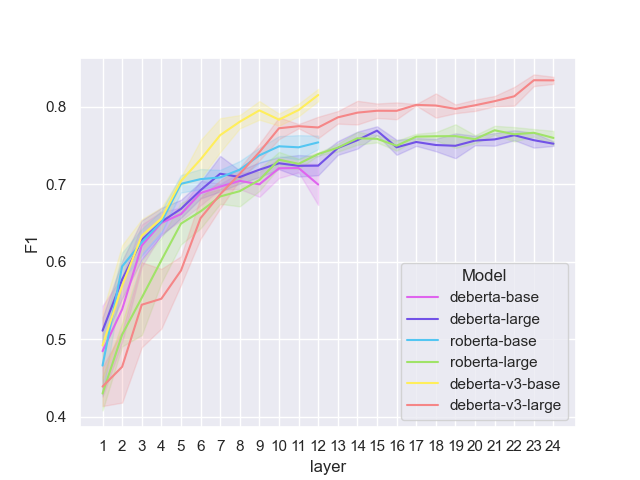}
         \caption{CoLA}
         \label{fig:layers_cola}
         
     \end{subfigure}
    \caption{Layer-wise F1-scores (average and standard deviation) for DeBERTa original and v3 \emph{large} and \emph{base} and RoBERTa \emph{large} and \emph{base} across 5 runs and 7 tasks ((a) Trustpilot Age to (g) CoLA).}
    \label{fig:layers}
\end{figure*}
\begin{figure*}[th!]
     \centering
     \begin{subfigure}[b]{0.248\textwidth}
         \centering
         \includegraphics[width=\textwidth, trim=2em 0.3em 2em 0.3em, clip]{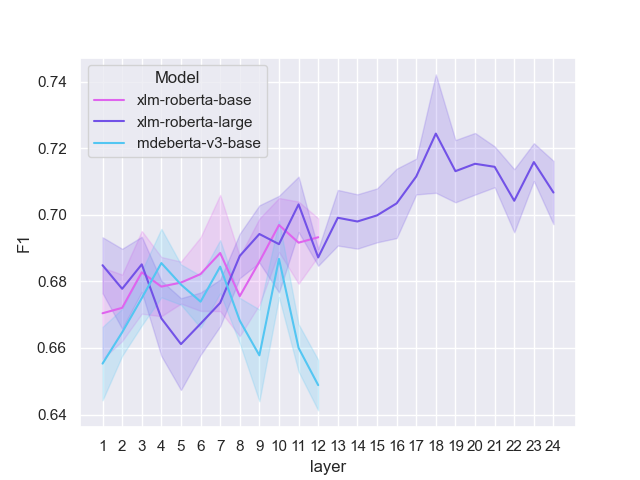}
         \caption{Trustpilot (Age)}
         \label{fig:layers_tp_age_m}
     \end{subfigure}
     \begin{subfigure}[b]{0.248\textwidth}
         \centering
         \includegraphics[width=\textwidth, trim=2em 0.3em 2em 0.3em, clip]{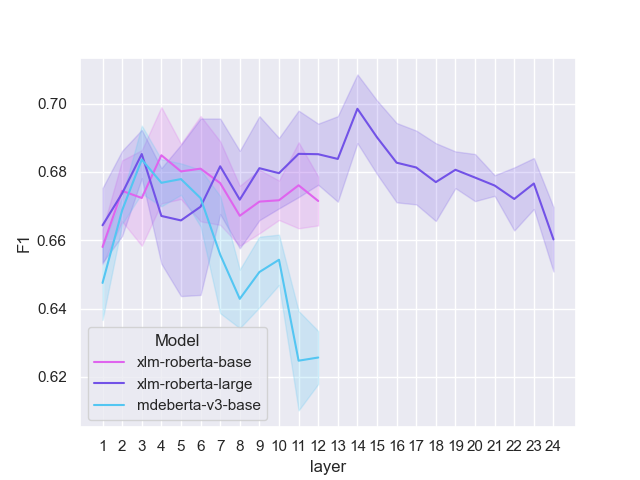}
         \caption{Trustpilot (Gender)}
         \label{fig:layers_tp_gender_m}
     \end{subfigure}
          \begin{subfigure}[b]{0.248\textwidth}
         \centering
         \includegraphics[width=\textwidth, trim=2em 0.3em 2em 0.3em, clip]{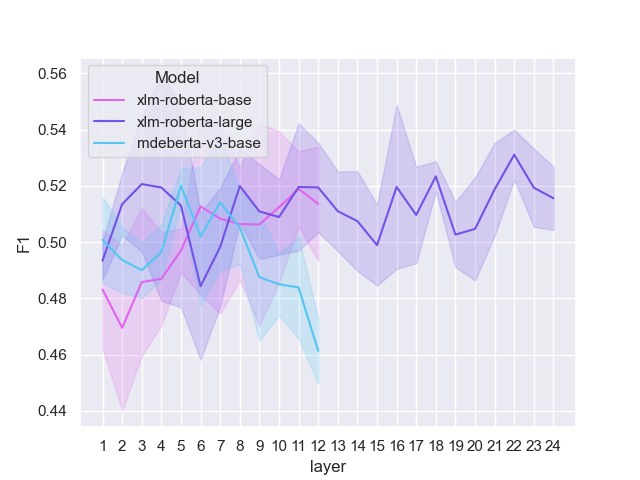}
         \caption{Facebook Congress}
         \label{fig:layers_fb_congress_m}
     \end{subfigure}
          \begin{subfigure}[b]{0.245\textwidth}
         \centering
         \includegraphics[width=\textwidth, trim=2em 0.3em 2em 0.3em, clip]{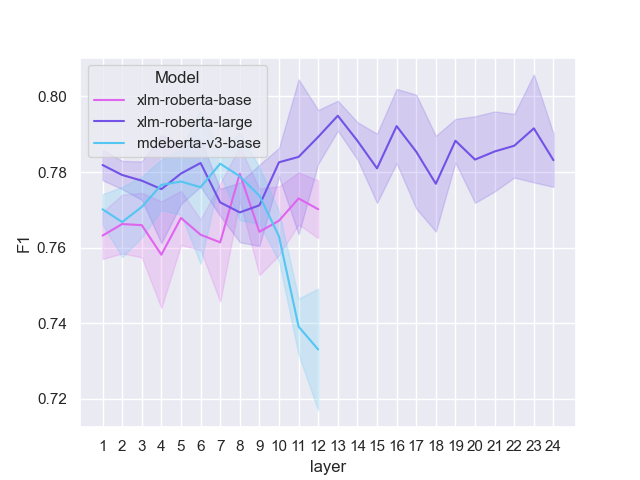}
         \caption{Facebook Wiki}
         \label{fig:layers_fb_wiki_m}
     \end{subfigure}
          \begin{subfigure}[b]{0.245\textwidth}
         \centering
         \includegraphics[width=\textwidth, trim=2em 0.3em 2em 0.3em, clip]{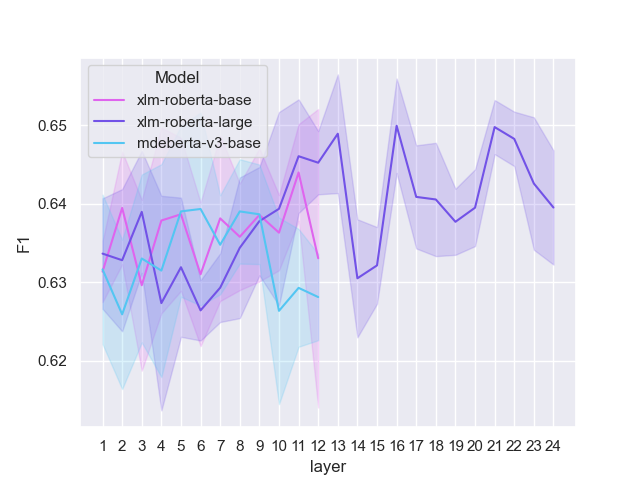}
         \caption{Fitocracy}
         \label{fig:layers_fito_m}
     \end{subfigure}
          \begin{subfigure}[b]{0.245\textwidth}
         \centering
         \includegraphics[width=\textwidth, trim=2em 0.3em 2em 0.3em, clip]{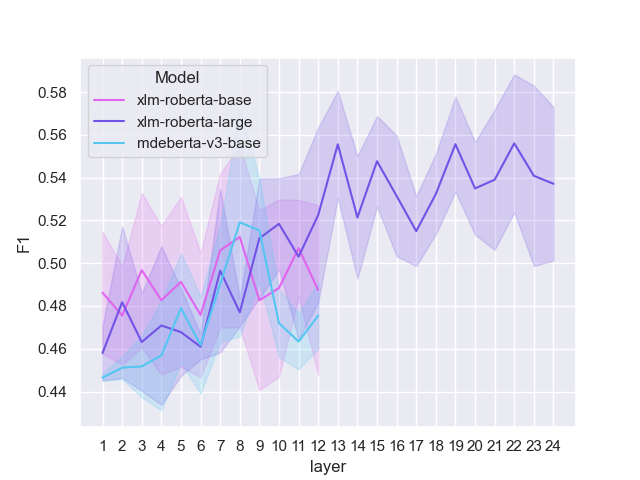}
         \caption{Reddit}
         \label{fig:layers_reddit_m}
     \end{subfigure}
          \begin{subfigure}[b]{0.245\textwidth}
         \centering
         \includegraphics[width=\textwidth, trim=2em 0.3em 2em 0.3em, clip]{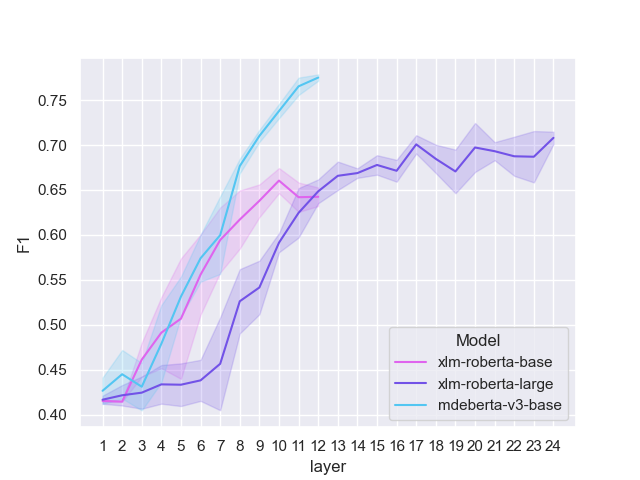}
         \caption{CoLA}
         \label{fig:layers_cola_m}
     \end{subfigure}
    \caption{Results for our analysis of multilingual models (RQ4). We show  F1-scores (average and standard deviation) across 5 runs on 7 tasks ((a) Trustpilot (Age) to (g) CoLA). The features we probe are extracted from different layers of XLM-RoBERTa \emph{large}, XLM-RoBERTa \emph{base}, and mDeBERTa \emph{base}.}
    \label{fig:layers_m}
\end{figure*}
\begin{figure*}[th!]
     \centering
     \begin{subfigure}[b]{0.248\textwidth}
         \centering
         \includegraphics[width=\textwidth, trim=2em 0.3em 2em 0.3em, clip]{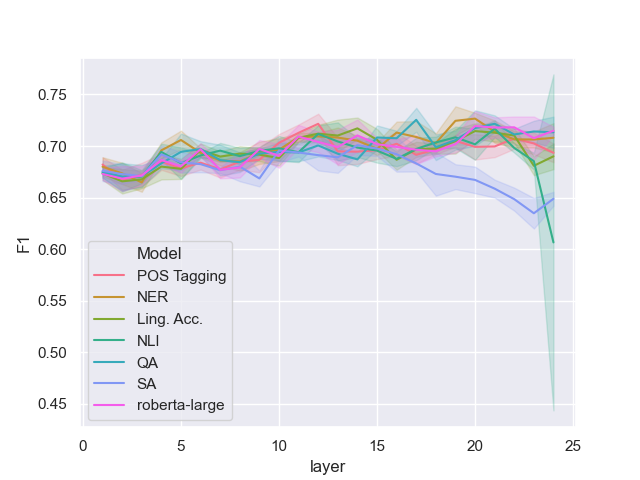}
         \caption{Trustpilot (Age)}
         \label{fig:stilt_tp_age}
     \end{subfigure}
     \begin{subfigure}[b]{0.248\textwidth}
         \centering
         \includegraphics[width=\textwidth, trim=2em 0.3em 2em 0.3em, clip]{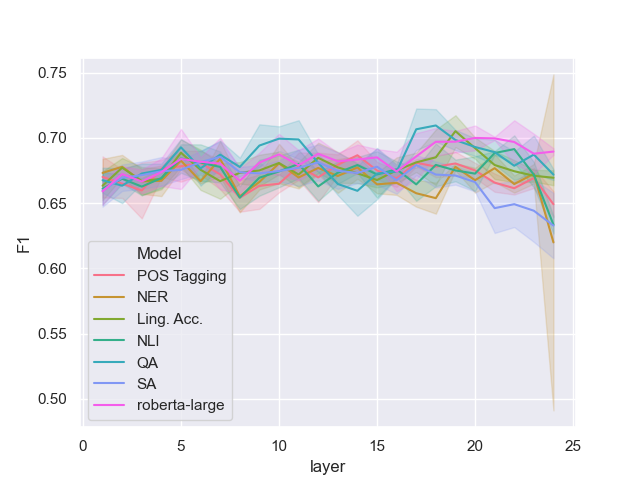}
         \caption{Trustpilot (Gender)}
         \label{fig:stilt_tp_gender}
     \end{subfigure}
          \begin{subfigure}[b]{0.248\textwidth}
         \centering
         \includegraphics[width=\textwidth, trim=2em 0.3em 2em 0.3em, clip]{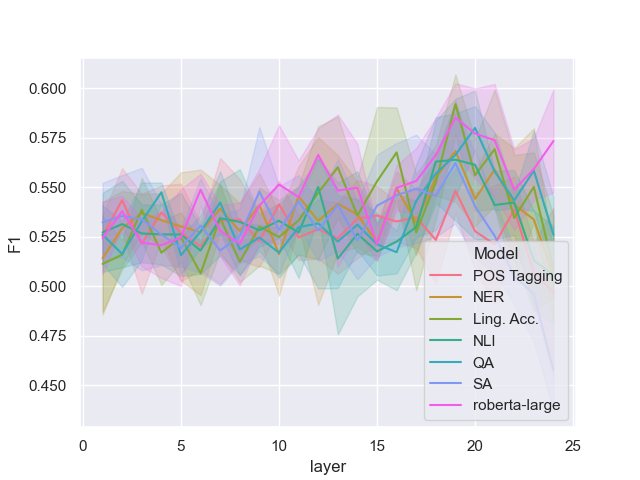}
         \caption{Facebook Congress}
         \label{fig:stilt_fb_congress}
     \end{subfigure}
          \begin{subfigure}[b]{0.245\textwidth}
         \centering
         \includegraphics[width=\textwidth, trim=2em 0.3em 2em 0.3em, clip]{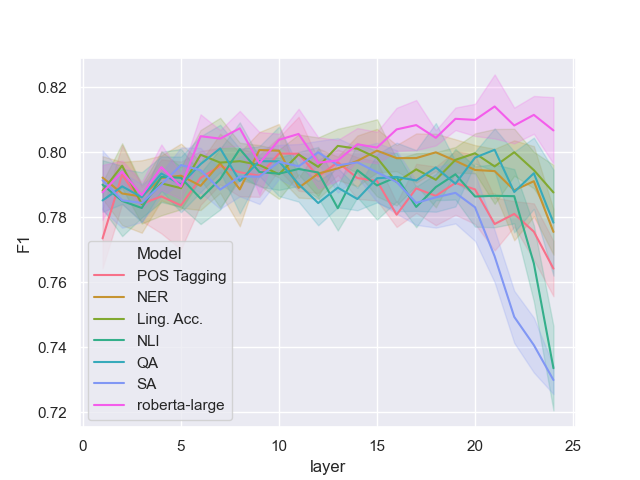}
         \caption{Facebook Wiki}
         \label{fig:stilt_fb_wiki}
     \end{subfigure}
          \begin{subfigure}[b]{0.245\textwidth}
         \centering
         \includegraphics[width=\textwidth, trim=2em 0.3em 2em 0.3em, clip]{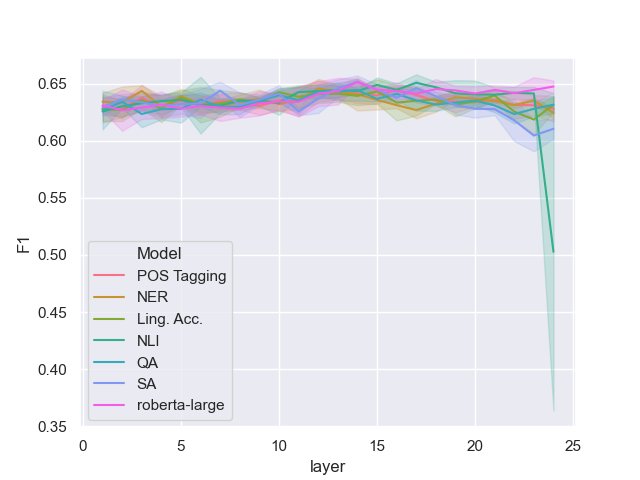}
         \caption{Fitocracy}
         \label{fig:stilt_fito}
     \end{subfigure}
          \begin{subfigure}[b]{0.245\textwidth}
         \centering
         \includegraphics[width=\textwidth, trim=2em 0.3em 2em 0.3em, clip]{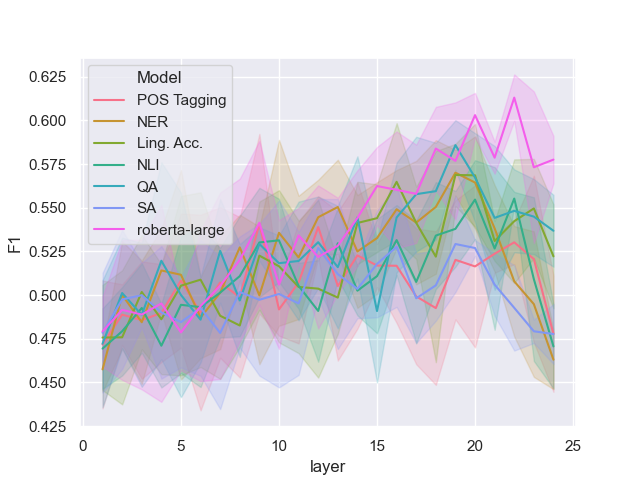}
         \caption{Reddit}
         \label{fig:stilt_reddit}
     \end{subfigure}
          \begin{subfigure}[b]{0.245\textwidth}
         \centering
         \includegraphics[width=\textwidth, trim=2em 0.3em 2em 0.3em, clip]{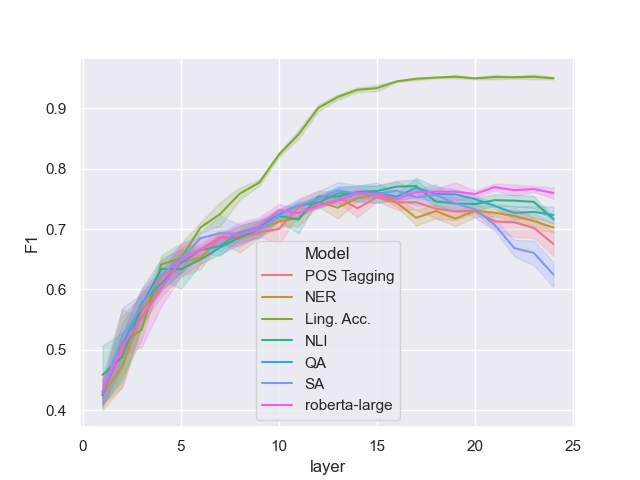}
         \caption{CoLA}
         \label{fig:stilt_cola}
     \end{subfigure}
    \caption{\al{Results of our STILTs analysis (RQ5) in terms of F1 scores (average and standard deviation) for 5 runs across 7 tasks ((a) Trustpilot Age to (g) CoLA). The features we probe are extracted from the original RoBERTa \emph{large}, and RoBERTa \emph{large} models, which we obtain from fine-tuning RoBERTa on 6 tasks: Part-of-Speech Tagging (POS Tagging), Named Entity Recognition (NER), Linguistic Acceptability Prediction (Ling. Acc.), Natural Language Inference (NLI), Question Answering (QA), and Sentiment Analysis (SA).}}
    \label{fig:stilt}
\end{figure*}

\paragraph{Results.} 
\al{The results for the classic and MDL probing are depicted in Figures \ref{fig:miniberta_classic} and \ref{fig:miniberta_online}, respectively.} Across all data sets, the sociodemographic classification improves with more pretraining data. The learning curves \al{in the classic probing} do not flatten out, which indicates the potential for more research on the topic. We conclude that \emph{with more pretraining data, more sociodemographic knowledge is present in the features.} This finding contrasts with other lower-level tasks, such as syntactic knowledge ~\citep{zhang-etal-2021-need, perez2021much}. As we hypothesized, though, it is similar to other higher-level language aspects, like common sense knowledge. Our control task, CoLA, exactly reflects this trend: the learning curve of predicting linguistic acceptability is much steeper. %

\paragraph{Cost-benefit Analysis.} Inspired by \citet{perez2021much}, we conduct a cost-benefit analysis. The authors base their estimate on the  costs provided in \citet{strubell-etal-2019-energy}. We follow their approach,\footnote{\al{Note, that these are presumably a overestimates, as hardware has been getting cheaper and more power efficient.}} and %
approximate the financial costs of training a model with \$60,948 / 30B words * \#TrainingWords for each of the MiniBERTas, and the CO$_2$ emissions of each MiniBERTa model as 6,990 lbs / 30B * \#TrainingWords. The final cost needs to be scaled with the number of pre-training procedures needed for model optimization reported by \citet{warstadt-etal-2020-learning}  (10 times for the 1B MiniBERTa, 25 times for the other MiniBERTa models). In contrast to \citet{perez2021much}, we also include RoBERTa \emph{base} in our analysis and scale the costs accordingly. 
Table~\ref{tab:costs} shows the cost estimates and expected performance improvements. Between 1M and 1B tokens the expected gains flatten (see previous analysis), while the gains are lower than the ones reported by \citet{perez2021much} for syntactic tasks. However, with 30B we can expect a large performance improvement indicating that higher performance can only be expected at even higher financial and environmental costs. Given that the already high baseline costs, such a development is ethically problematic. Our results support the  need for more research on sustainable NLP, \al{especially when tasks require in-depth language understanding}.

\subsection{RQ3: Where is sociodemographic knowledge located?}
We test embeddings extracted from different layers.

\paragraph{Approach.} 
In the previous experiments, we followed the standard approach and pooled representations from the last layer of the Transformer. In contrast, here we test the average pooled representations from \textit{each} layer $n \in [1:\textnormal{num\_layers}]$, where $\textnormal{num\_layers}$ corresponds to the number of layers in the model. We test RoBERTa and DeBERTa (original and v3) in the \emph{large} and \emph{base} configurations.

\paragraph{Results.}
We show the results in Figures \ref{fig:layers_tp_age}--\ref{fig:layers_cola}. Note the relatively high standard deviations compared to the overall performance range. The  exception to this observation is again CoLA, our control task  (Figure~\ref{fig:layers_cola}). 
The tendency seems to be that higher layers offer better representations for sociodemographic classification (e.g., Trustpilot (Age) in Figure~\ref{fig:layers_tp_age}, Reddit (Gender) in Figure~\ref{fig:layers_reddit}), but performance improvement across layers is much more skewed for CoLA than for the sociodemographic probing tasks. Especially for DeBERTa v3 \emph{base}, the probing results are often better for lower model layers (e.g., Figure~\ref{fig:layers_congress}). 
This runs counter to \citet{tenney-etal-2019-bert}, who showed higher-level semantic knowledge to be encoded in the higher layers of BERT.  We conclude that \emph{sociodemographic knowledge is much less localized} in PLMs than lower-level knowledge. This finding corresponds to the observation that different sociodemographic factors are expressed in different ways \cite{johannsen-etal-2015-cross}.
As in our experiments for answering RQ1, DeBERTa \emph{large} v3 has superior knowledge about lower-level linguistic aspects, but not sociodemographic knowledge.

\subsection{RQ4: Does the sociodemographic knowledge in multilingual models differ?}
\al{\citet{hung2022can} recently showed that  straight-forward attempts to (socio)demographic adaptation of multilingual Transformers can lead to a better separation of representation areas according to input text languages and not according to author demographics. This leads us to question whether the multilingual signal significantly affects the encoding of sociodemographic knowledge in the models. 
We probe multilingual PLMs for their encoding of sociodemographics in English and further validate our previous findings}.

\paragraph{Approach.} 
We use multilingual versions of RoBERTa and DeBERTa available on Huggingface: XLM-RoBERTa in \emph{large} and \emph{base} configuration and mDeBERTa v3 in \emph{base} configuration.\footnote{No \emph{large} configuration of mDeBERTa was available}

\paragraph{Result.} 
We only show the classic probing results (Figure~\ref{fig:layers_m}, see Appendix for MDL). 
While the scores are slightly lower than for monolingual PLMs, they are generally in-line with our findings from RQ2: sociodemographic knowledge is less localized than that for CoLA. While for XLM-RoBERTa \emph{large} and \emph{base} higher layers encode the sociodemographics, the results of DeBERTa v3 \emph{base} show an opposite trend. We conclude that \emph{the localization of sociodemographic knowledge in multilingual models follows their monolingual counterparts}. The layer-wise behavior of DeBERTa v3 \emph{base} is an additional pointer to the effect of the training regimes on the sociodemographic knowledge encoded in PLMs.

\subsection{RQ5: What is the effect of STILTs on the encoding of the knowledge?}
We explore the effect of STILTs on the sociodemographic knowledge encoded in the layers.

\paragraph{Approach.} We use the encoders of readily fine-tuned RoBERTa \emph{large} models from the Huggingface repository trained on the following tasks and data sets: POS tagging and dependency parsing on UPOS, named entity recognition, natural language inference on MNLI, question answering on SQuaD v.2., sentiment analysis on SST2, and linguistic acceptability prediction on our control task CoLA.

\paragraph{Results.} 
See Figure~\ref{fig:stilt} for the classic probing results (MDL probing results in the Appendix). 
Unsurprisingly, supplementary training on Linguistic Acceptability Prediction (=our control task CoLA), leads to superior representations for CoLA probing (Figure~\ref{fig:stilt_cola}). This effect is clearly visible from layer 5 onwards, indicating that the top 19 layers become specialized during the STILTs fine-tuning. In contrast, the selected STILTs tasks do not improve the sociodemographic knowledge in the  representations (e.g., QA STILTs for gender prediction in Trustpilot) or even reduce that knowledge (e.g., NLI and SA STILTs for gender prediction in FB Wiki (Figure~\ref{fig:stilt_fb_wiki})). 
The results suggest that sociodemographic knowledge is  overwritten during STILTs. Interestingly, this effect mostly occurs on the last 5 to 10 layers  (e.g., Trustpilot Age prediction from layer 10 (Figure~\ref{fig:stilt_tp_age}), and much more gently than the CoLA improvement.

%% file: 06-conclusion.tex
Sociodemographic aspects shape our language and are thus important factors to model in language technology. However, despite a plethora of works probing PLMs for various types of knowledge, we know little about these higher-level aspects of language. We present \suite{}  to understand \emph{whether}, \emph{when}, and \emph{where} PLMs encode sociodemographic knowledge in their representations. 
We find that sociodemograophic knowledge is located in PLMs, but much more diffuse than lower-level aspects. \al{In the future, we will extend our analysis to  languages other than English.} 
We hope that  our findings will fuel further research towards human-like language understanding. %

%% file: 07-limitations.tex
Our work deals with predicting sociodemographic aspects from text, which should be considered sensitive information. Predictive methods can result in potentially harmful applications, e.g., in the context of user profiling. We acknowledge this potential for \emph{dual use}~\citep{jonas1984imperative} of the data sets we use. However, in this work, we are interested in advancing NLP research towards a \emph{better understanding of such fine-grained aspects of language and how they are already captured by our technology}. We believe that these insights will lead us toward fairer and more inclusive language technology. 
In contrast, we explicitly discourage the prediction of sensitive attributes from text for harmful purposes. 

Further, we acknowledge that our work is limited in that the data sets available to us model gender as a binary variable, which does not reflect the wide variety of possible identites along the gender spectrum and beyond~\citep{lauscher2022welcome}. However, we are not aware of other suitable data sets without this limitation. We have reason to believe, though, that even the findings derived from a binary view on gender (as well as for age) can provide an initial understanding of how language varies, and that any results will hold under a more sophisticated modeling of the problem.

An additional limitation of our work comes from the pre-trained models we used. All the models tested are easily-downloadable single-GPU models that have been pre-trained on general-purpose data. We acknowledge that results might differ for  models that were of bigger capacity and  pre-trained on data from other and more specific domains, e.g., social media. 
The same argument can be made about the architectures used. We mainly focused on BERT-like models trained via MLM, which are only a subset of the language models proposed in the literature. We leave the exploration of these effects (e.g., pre-training objective) for future work.

%% file: xx-appendix.tex
\label{sec:appendix}
\section{Models Used}
We provide an overview on all models we have used in this study. They are available on the Huggingface Hub: \url{https://huggingface.co}.

\subsection{Models Used for RQ1}
For our base experiment we have used the following models.
\begin{itemize}
     \item \texttt{roberta-base}: 12 layers, 12 attention heads, hidden size of 768
 \item \texttt{roberta-large}: 24 layers, 16 attention heads, hidden size of 1024
   \item \texttt{microsoft/deberta-v3-base}: 12 layers, 12 attention heads, hidden size of 768
  \item \texttt{microsoft/deberta-v3-large}: 24 layers, 16 attention heads, hidden size of 1024
  \item \texttt{microsoft/deberta-v3-xsmall}: 12 layers, 6 attention heads, hidden size of 384
  \item \texttt{microsoft/deberta-v3-small}: 6 layers, 12 attention heads, hidden size of 768
     \item \texttt{microsoft/deberta-base}: 12 layers, 12 attention heads, hidden size of 768
  \item \texttt{microsoft/deberta-large}: 24 layers, 16 attention heads, hidden size of 1024
\end{itemize}

\subsection{Models Used for RQ2}
For investigating the amount of pre-training data needed, we have used the suite of MiniBERTas, and RoBERTa \emph{base}.
\begin{itemize}
     \item \texttt{roberta-base}: 12 layers, 12 attention heads, hidden size of 768
\item \texttt{nyu-mll/roberta-base-1B-1}: 12 layers, 12 attention heads, hidden size of 768
\item \texttt{nyu-mll/roberta-base-1B-2}: 12 layers, 12 attention heads, hidden size of 768
\item \texttt{nyu-mll/roberta-base-1B-3}: 12 layers, 12 attention heads, hidden size of 768
\item \texttt{nyu-mll/roberta-base-100M-1}: 12 layers, 12 attention heads, hidden size of 768
\item \texttt{nyu-mll/roberta-base-100M-2}: 12 layers, 12 attention heads, hidden size of 768
\item \texttt{nyu-mll/roberta-base-100M-3}: 12 layers, 12 attention heads, hidden size of 768
\item \texttt{nyu-mll/roberta-base-10M-1}: 12 layers, 12 attention heads, hidden size of 768
 \item \texttt{nyu-mll/roberta-base-10M-2}: 12 layers, 12 attention heads, hidden size of 768
 \item \texttt{nyu-mll/roberta-base-10M-3}: 12 layers, 12 attention heads, hidden size of 768
 \item \texttt{nyu-mll/roberta-med-small-1M-1}: 6 layers, 8 attention heads, hidden size of 512
 \item \texttt{nyu-mll/roberta-med-small-1M-2}: 6 layers, 8 attention heads, hidden size of 512
 \item \texttt{nyu-mll/roberta-med-small-1M-3}: 6 layers, 8 attention heads, hidden size of 512
\end{itemize}

\subsection{Models Used for RQ3}
We investigated the layer-wise knowledge of the following models.
\begin{itemize}
     \item \texttt{roberta-base}: 12 layers, 12 attention heads, hidden size of 768
 \item \texttt{roberta-large}: 24 layers, 16 attention heads, hidden size of 1024
   \item \texttt{microsoft/deberta-v3-base}: 12 layers, 12 attention heads, hidden size of 768
  \item \texttt{microsoft/deberta-v3-large}: 24 layers, 16 attention heads, hidden size of 1024
     \item \texttt{microsoft/deberta-base}: 12 layers, 12 attention heads, hidden size of 768
  \item \texttt{microsoft/deberta-large}: 24 layers, 16 attention heads, hidden size of 1024
\end{itemize}

\subsection{Models Used for RQ4}
As multilingual counter-parts, we employed the following models.
\begin{itemize}
     \item \texttt{xlm-roberta-base}: 12 layers, 12 attention heads, hidden size of 768
 \item \texttt{xlm-roberta-large}: 24 layers, 16 heads, hidden size of 1024
   \item \texttt{microsoft/mdeberta-v3-base}: 12 layers, 12 attention heads, hidden size of 768
\end{itemize}

\subsection{Models Used for RQ5}
Finally, we ran the STILT experiment with the following models.
\begin{itemize}
\item \texttt{roberta-large}: 24 layers, 16 heads, hidden size of 1024
 \item \texttt{KoichiYasuoka/roberta-large} \texttt{-english-upos}: 24 layers, 16 heads, hidden size of 1024
\item \texttt{Jean-Baptiste/roberta-large} \texttt{-ner-english}: 24 layers, 16 heads, hidden size of 1024
\item \texttt{cointegrated/roberta-large} \texttt{-cola-krishna2020}: 24 layers, 16 heads, hidden size of 1024
\item \texttt{roberta-large-mnli}: 24 layers, 16 heads, hidden size of 1024
\item \texttt{navteca/roberta-large-squad2}: 24 layers, 16 heads, hidden size of 1024
\item \texttt{howey/roberta-large-sst2}: 24 layers, 16 heads, hidden size of 1024
\end{itemize}

\section{Additional Results}
We provide the additional results for MDL probing.

\subsection{Additional Results for RQ3}
\begin{figure*}[t]
     \centering
     \begin{subfigure}[b]{0.248\textwidth}
         \centering
         \includegraphics[width=\textwidth, trim=2em 0.3em 2em 0.3em, clip]{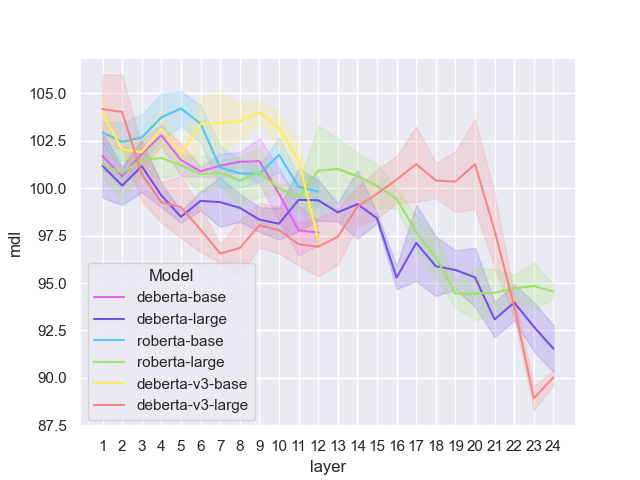}
         \caption{Trustpilot (Age)}
         \label{fig:layers_tp_age_o}
     \end{subfigure}
     \begin{subfigure}[b]{0.248\textwidth}
         \centering
         \includegraphics[width=\textwidth, trim=2em 0.3em 2em 0.3em, clip]{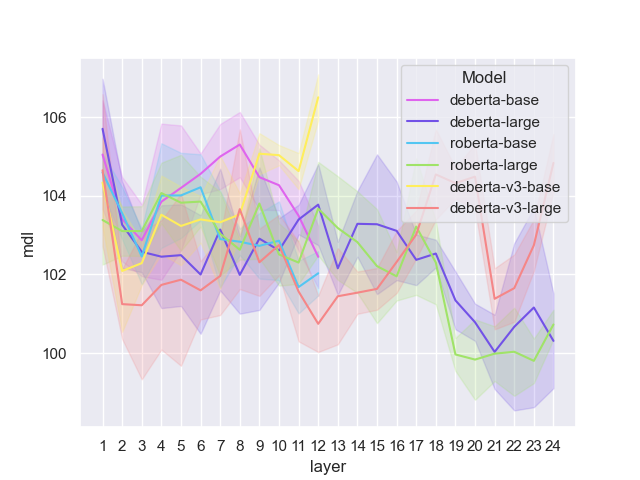}
         \caption{Trustpilot (Gender)}
         \label{fig:layers_tp_gender_o}
     \end{subfigure}
          \begin{subfigure}[b]{0.248\textwidth}
         \centering
         \includegraphics[width=\textwidth, trim=2em 0.3em 2em 0.3em, clip]{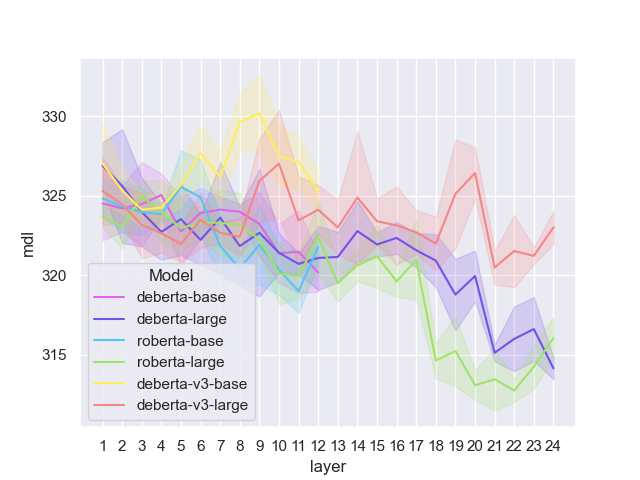}
         \caption{FB Congress (Gender)}
         \label{fig:layers_congress_o}
     \end{subfigure}
          \begin{subfigure}[b]{0.245\textwidth}
         \centering
         \includegraphics[width=\textwidth, trim=2em 0.3em 2em 0.3em, clip]{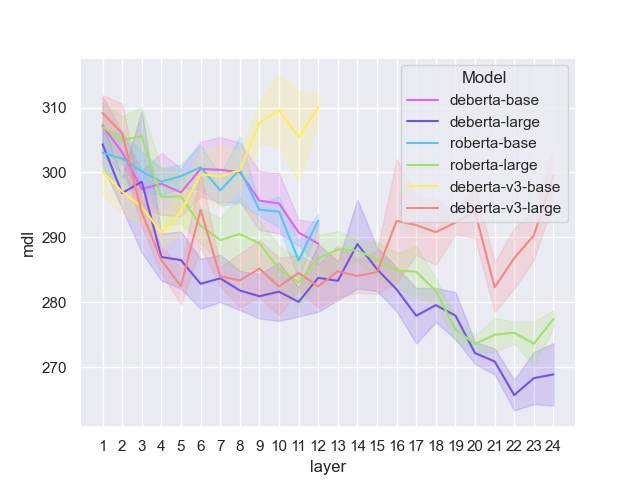}
         \caption{FB Public (Gender)}
         \label{fig:layers_wiki_o}
     \end{subfigure}
          \begin{subfigure}[b]{0.245\textwidth}
         \centering
         \includegraphics[width=\textwidth, trim=2em 0.3em 2em 0.3em, clip]{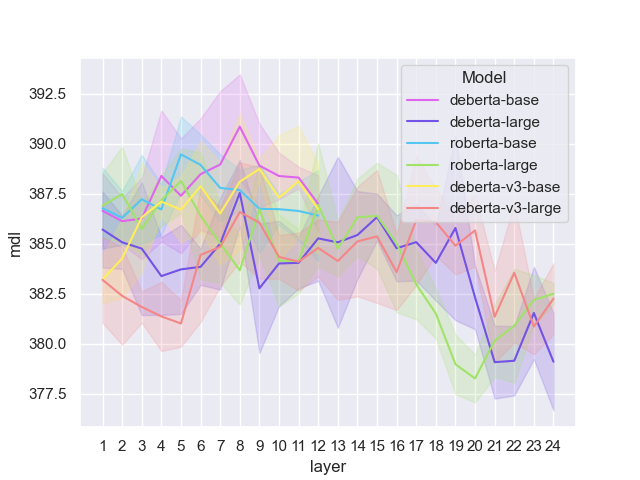}
         \caption{Fitocracy (Gender)}
         \label{fig:layers_fitocracy_o}
     \end{subfigure}
          \begin{subfigure}[b]{0.245\textwidth}
         \centering
         \includegraphics[width=\textwidth, trim=2em 0.3em 2em 0.3em, clip]{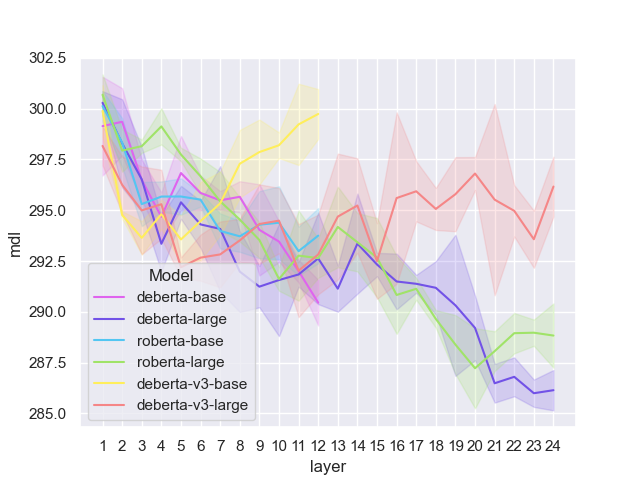}
         \caption{Reddit (Gender)}
         \label{fig:layers_reddit_o}
         
     \end{subfigure}
          \begin{subfigure}[b]{0.245\textwidth}
         \centering
         \includegraphics[width=\textwidth, trim=2em 0.3em 2em 0.3em, clip]{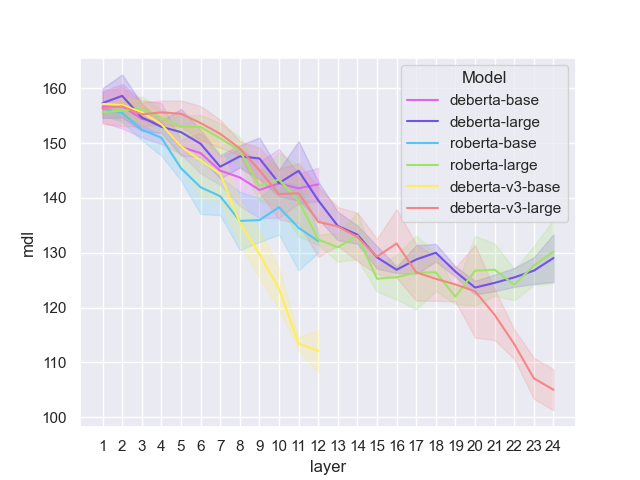}
         \caption{CoLA}
         \label{fig:layers_cola_o}
         
     \end{subfigure}
    \caption{Results showing our layer-wise analysis of DeBERTa original and v3 \emph{large} and \emph{base} and RoBERTa \emph{large} and \emph{base} in terms of average and standard deviation of the MDL for 5 runs across 7 tasks ((a) Trustpilot Age to (g) CoLA)}
    \label{fig:layers_o}
\end{figure*}
The layer-wise analysis for MDL probing is provided in Figure~\ref{fig:layers_o}.

\subsection{Additional Results for RQ4}

\begin{figure*}[th!]
     \centering
     \begin{subfigure}[b]{0.248\textwidth}
         \centering
         \includegraphics[width=\textwidth, trim=2em 0.3em 2em 0.3em, clip]{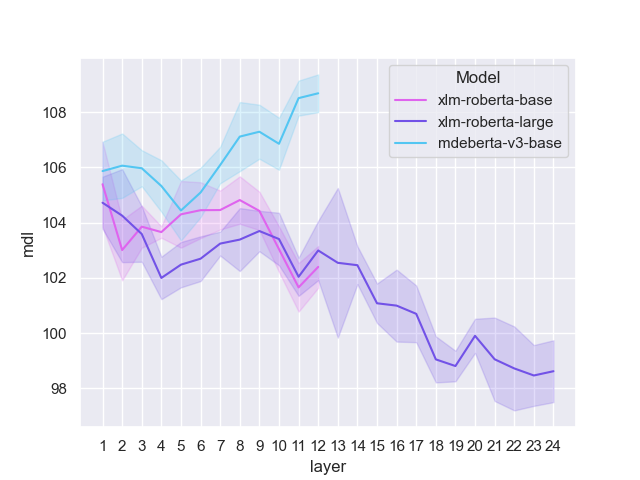}
         \caption{Trustpilot (Age)}
     \end{subfigure}
     \begin{subfigure}[b]{0.248\textwidth}
         \centering
         \includegraphics[width=\textwidth, trim=2em 0.3em 2em 0.3em, clip]{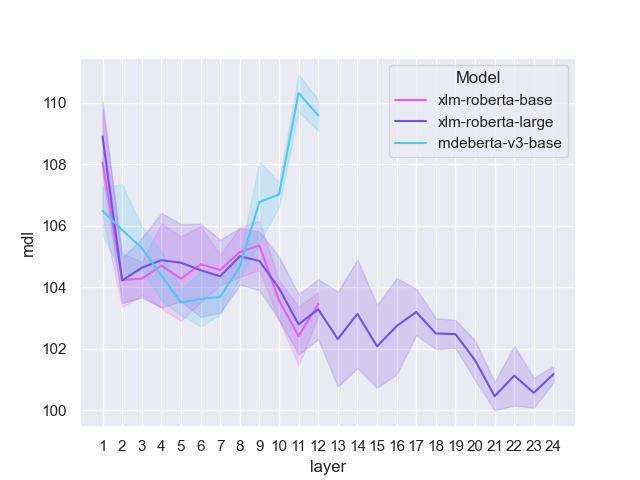}
         \caption{Trustpilot (Gender)}
     \end{subfigure}
          \begin{subfigure}[b]{0.248\textwidth}
         \centering
         \includegraphics[width=\textwidth, trim=2em 0.3em 2em 0.3em, clip]{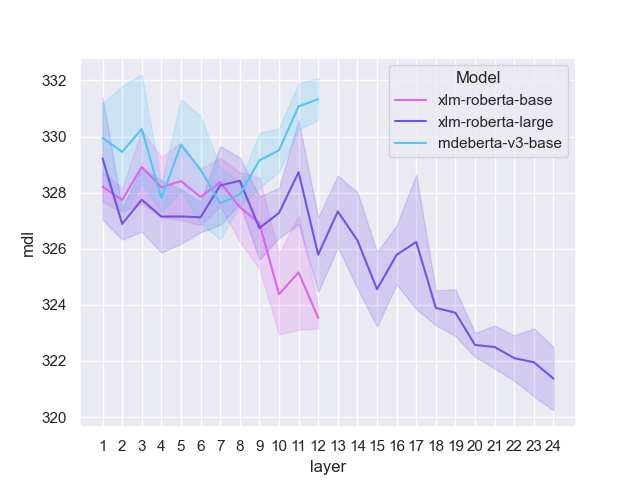}
         \caption{FB Congress (Gender)}
     \end{subfigure}
          \begin{subfigure}[b]{0.245\textwidth}
         \centering
         \includegraphics[width=\textwidth, trim=2em 0.3em 2em 0.3em, clip]{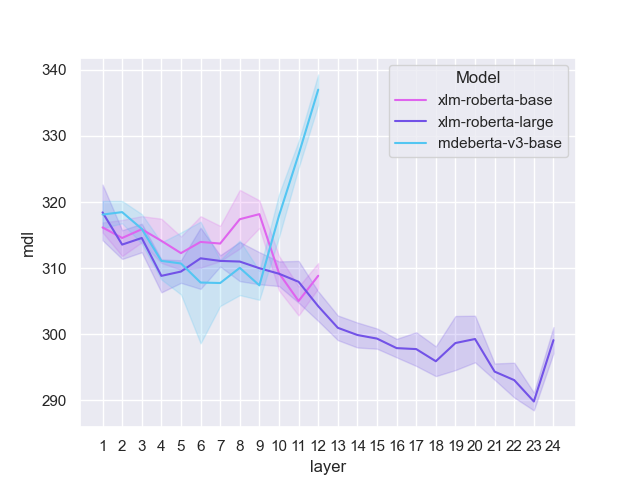}
         \caption{FB Public (Gender)}
     \end{subfigure}
          \begin{subfigure}[b]{0.245\textwidth}
         \centering
         \includegraphics[width=\textwidth, trim=2em 0.3em 2em 0.3em, clip]{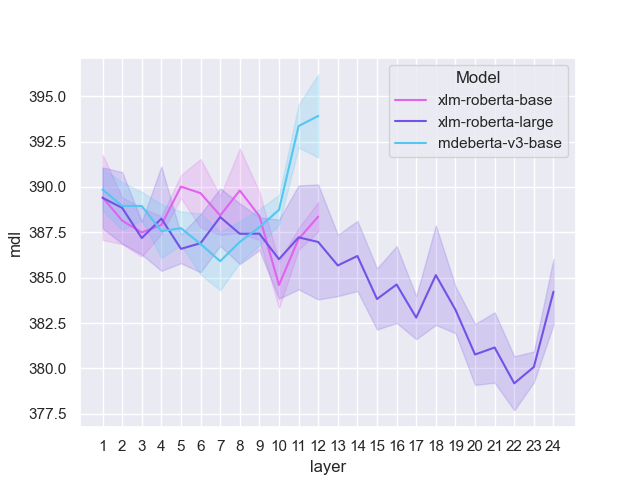}
         \caption{Fitocracy (Gender)}
     \end{subfigure}
          \begin{subfigure}[b]{0.245\textwidth}
         \centering
         \includegraphics[width=\textwidth, trim=2em 0.3em 2em 0.3em, clip]{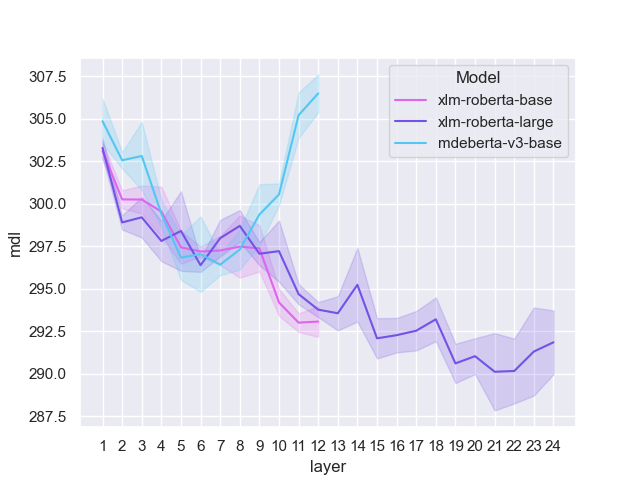}
         \caption{Reddit (Gender)}
     \end{subfigure}
          \begin{subfigure}[b]{0.245\textwidth}
         \centering
         \includegraphics[width=\textwidth, trim=2em 0.3em 2em 0.3em, clip]{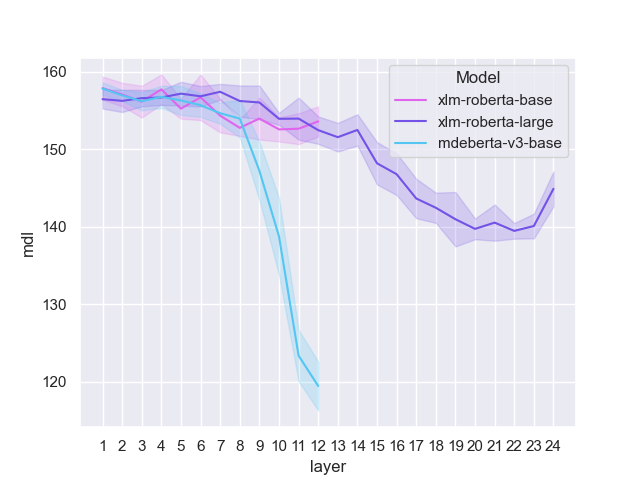}
         \caption{CoLA}
     \end{subfigure}
    \caption{Results for our analysis of multilingual models in terms of average and standard deviation of the MDL scores for 5 runs across 7 tasks ((a) Trustpilot (Age) to (g) CoLA) for features extracted from different layers of XLM-RoBERTa \emph{large} and \emph{base} and mDeBERTa \emph{base}.}
    \label{fig:layers_m_o}
\end{figure*}

We show the MDL results for the multilingual analysis in Figure~\ref{fig:layers_m_o}.

\subsection{Additional Results for RQ5}
\begin{figure*}[th!]
     \centering
     \begin{subfigure}[b]{0.248\textwidth}
         \centering
         \includegraphics[width=\textwidth, trim=2em 0.3em 2em 0.3em, clip]{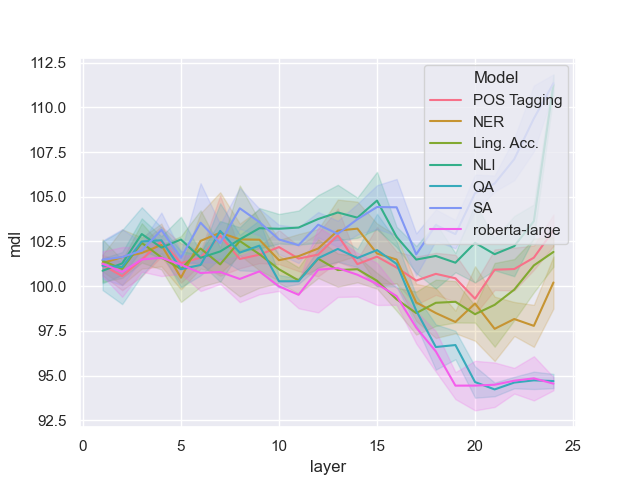}
         \caption{Trustpilot (Age)}
         \label{fig:stilt_tp_age_online}
     \end{subfigure}
     \begin{subfigure}[b]{0.248\textwidth}
         \centering
         \includegraphics[width=\textwidth, trim=2em 0.3em 2em 0.3em, clip]{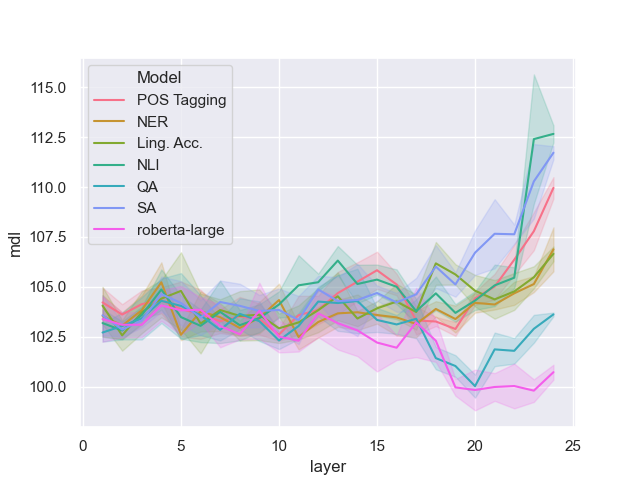}
         \caption{Trustpilot (Gender)}
         \label{fig:stilt_tp_gender_online}
     \end{subfigure}
          \begin{subfigure}[b]{0.248\textwidth}
         \centering
         \includegraphics[width=\textwidth, trim=2em 0.3em 2em 0.3em, clip]{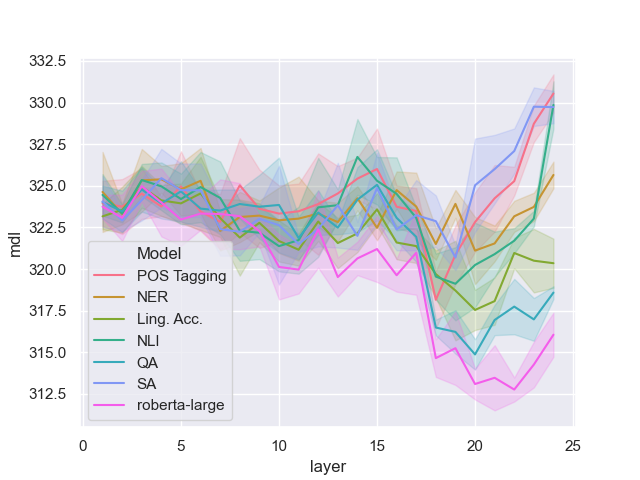}
         \caption{FB Congress (Gender)}
         \label{fig:stilt_fb_congress_online}
     \end{subfigure}
          \begin{subfigure}[b]{0.245\textwidth}
         \centering
         \includegraphics[width=\textwidth, trim=2em 0.3em 2em 0.3em, clip]{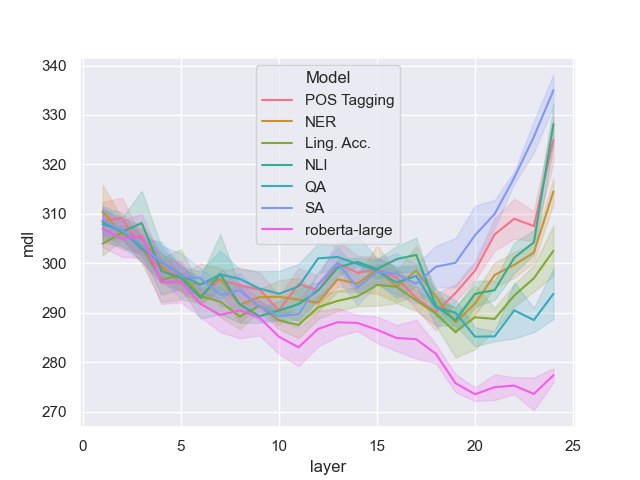}
         \caption{FB Public (Gender)}
         \label{fig:stilt_fb_wiki_online}
     \end{subfigure}
          \begin{subfigure}[b]{0.245\textwidth}
         \centering
         \includegraphics[width=\textwidth, trim=2em 0.3em 2em 0.3em, clip]{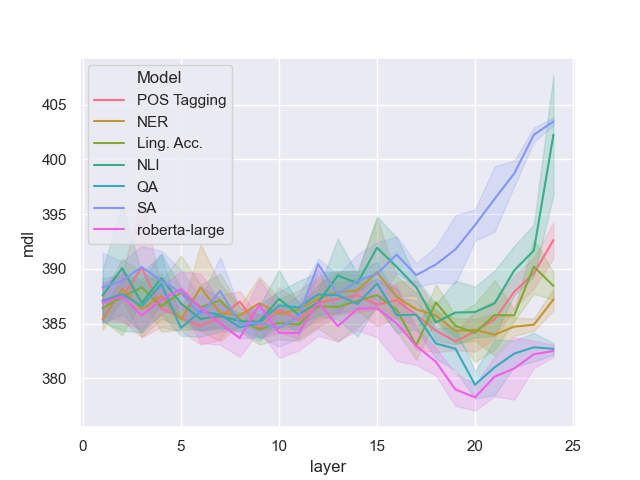}
         \caption{Fitocracy (Gender)}
         \label{fig:stilt_fito_online}
     \end{subfigure}
          \begin{subfigure}[b]{0.245\textwidth}
         \centering
         \includegraphics[width=\textwidth, trim=2em 0.3em 2em 0.3em, clip]{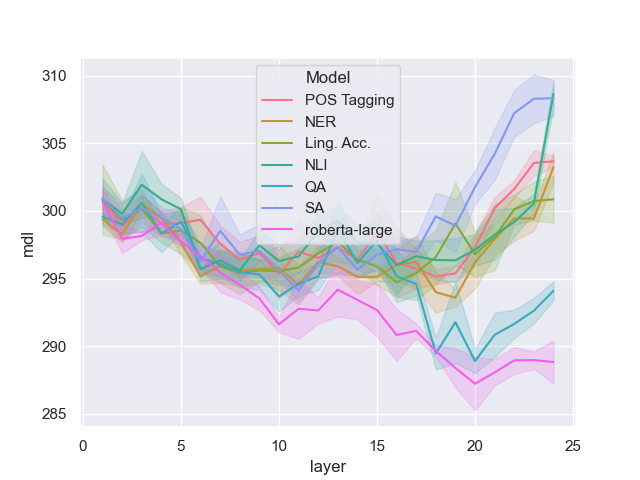}
         \caption{Reddit (Gender)}
         \label{fig:stilt_reddit_online}
     \end{subfigure}
          \begin{subfigure}[b]{0.245\textwidth}
         \centering
         \includegraphics[width=\textwidth, trim=2em 0.3em 2em 0.3em, clip]{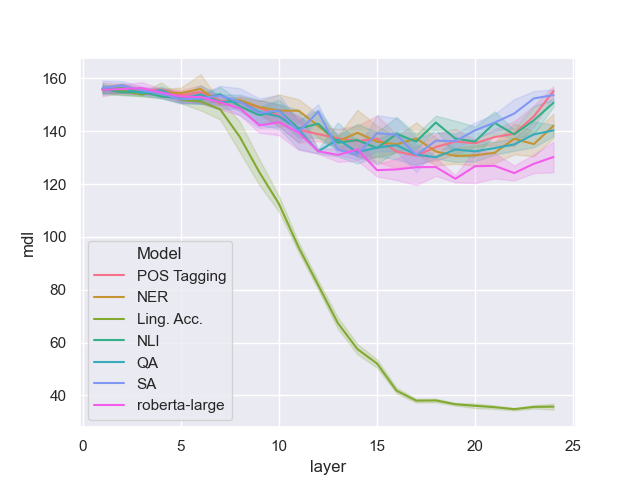}
         \caption{CoLA}
         \label{fig:stilt_cola_online}
     \end{subfigure}
    \caption{Results for our Supplementary Training on Intermediate Labeled Tasks (STILT) analysis. We show the probing results in terms of average and standard deviation of the MDL scores for 5 runs across 7 tasks ((a) Trustpilot Age to (g) CoLA) for features extracted from the original RoBERTa \emph{large}, and RoBERTa \emph{large} fine-tuned on 6 tasks: Part-of-Speech Tagging (POS Tagging), Named Entity Recognition (NER), Linguistic Acceptability Prediction (Ling. Acc.), Natural Language Inference (NLI), Question Answering (QA), and Sentiment Analysis (SA).}
    \label{fig:stilt_online}
\end{figure*}

We provide the MDL probing results for our STILT analysis in Figure~\ref{fig:stilt_online}.

